\documentclass[runningheads]{llncs}

\usepackage{eccv}

\usepackage{eccvabbrv}

\usepackage{graphicx}
\usepackage{amsmath}
\usepackage{amssymb}
\usepackage{booktabs}
\usepackage{arydshln}
\usepackage{xcolor}
\usepackage{caption}
\usepackage{multirow}
\usepackage{caption}
\usepackage{float}

\usepackage[accsupp]{axessibility}  

\usepackage{hyperref}

\usepackage{orcidlink}

\begin{document}

\title{SurgicaL-CD: Generating Surgical Images via Unpaired Image Translation with Latent Consistency Diffusion Models} 

\titlerunning{Surgical image generation with diffusion models}

\author{Danush Kumar Venkatesh\inst{1,2} \and
Dominik Rivoir\inst{1,3,*} \and
Micha Pfeiffer\inst{1,*} \and Stefanie Speidel\inst{1,2,3}}

\authorrunning{DK.~Venkatesh et al.}

\institute{NCT/UCC Dresden, Germany, DKFZ Heidelberg, Germany, Faculty of Medicine and University Hospital Carl Gustav Carus, TUD Dresden University of Technology, Dresden, Germany, HZDR Dresden, Germany \and 
Department of Translational Surgical Oncology, NCT/UCC Dresden, Faculty of Medicine and University Hospital Carl Gustav Carus, TUD Dresden University of Technology, Germany \and
The Centre for Tactile Internet with Human-in-the-Loop (CeTI), TUD Dresden University of Technology, Germany \\ * equal contribution \\
\email{\{danushkumar.venkatesh,dominik.rivor,micha.pfeiffer,\\stefanie.speidel\}@nct-dresden.de}}

\maketitle

\begin{abstract}
  Computer-assisted surgery (CAS) systems are designed to assist surgeons during procedures, thereby reducing complications and enhancing patient care. Training machine learning models for these systems requires a large corpus of annotated datasets, which is challenging to obtain in the surgical domain due to patient privacy concerns and the significant labeling effort required from doctors. Previous methods have explored unpaired image translation using generative models to create realistic surgical images from simulations. However, these approaches have struggled to produce high-quality, diverse surgical images. In this work, we introduce \emph{SurgicaL-CD}, a consistency-distilled diffusion method to generate realistic surgical images with only a few sampling steps without paired data. We evaluate our approach on three datasets, assessing the generated images in terms of quality and utility as downstream training datasets. Our results demonstrate that our method outperforms GANs and diffusion-based approaches. Our code is available at \url{https://gitlab.com/nct_tso_public/gan2diffusion}.
  \keywords{Surgical image generation \and Unpaired image translation \and Diffusion models}
\end{abstract}

\section{Introduction}
\label{sec:intro}
Significant breakthroughs in deep learning (DL) techniques have been achieved over the years, primarily driven by increased computational power and data availability~\cite{lecun2015deep}. These advancements have led to the emergence of Surgical Data Science (SDS)~\cite{maier2022surgical}, a nascent field aimed at improving interventional healthcare by analyzing patient data. For example, during surgery, assistive systems can examine real patient data and provide insights to surgeons, reducing complications and enhancing in-op and post-op patient care~\cite{bodenstedt2020artificial,haidegger2022robot,maier2017surgical}. A computational solution for such systems is training machine learning models to find and recognize patterns in image data. Despite the success of DL methods in other domains, their full potential in surgery has yet to be realized. A key obstacle is the requirement for large amounts of annotated data~\cite{maier2022surgical}.

The primary challenge is collecting surgical data, which involves patient consent, legal restrictions in surgical rooms~\cite{hager2020surgical} and data sharing, and procedural differences between the hospitals accounting for variability in datasets. This process is gradually being streamlined, with many surgical datasets being open-sourced for research~\cite{endonet,carstens2023dresden,nwoye2022rendezvous}. However, training large-scale models remains a bottleneck due to the need to annotate such datasets. In contrast to natural images, medical experts need to annotate them, and their time is very limited, which poses a significant practical challenge. Motivated by this challenge, we pose the research question: Can DL methods reduce the annotation effort by automatically developing labeled surgical datasets?

\begin{figure}[tb]
  \centering
  \includegraphics[height=6.5cm]{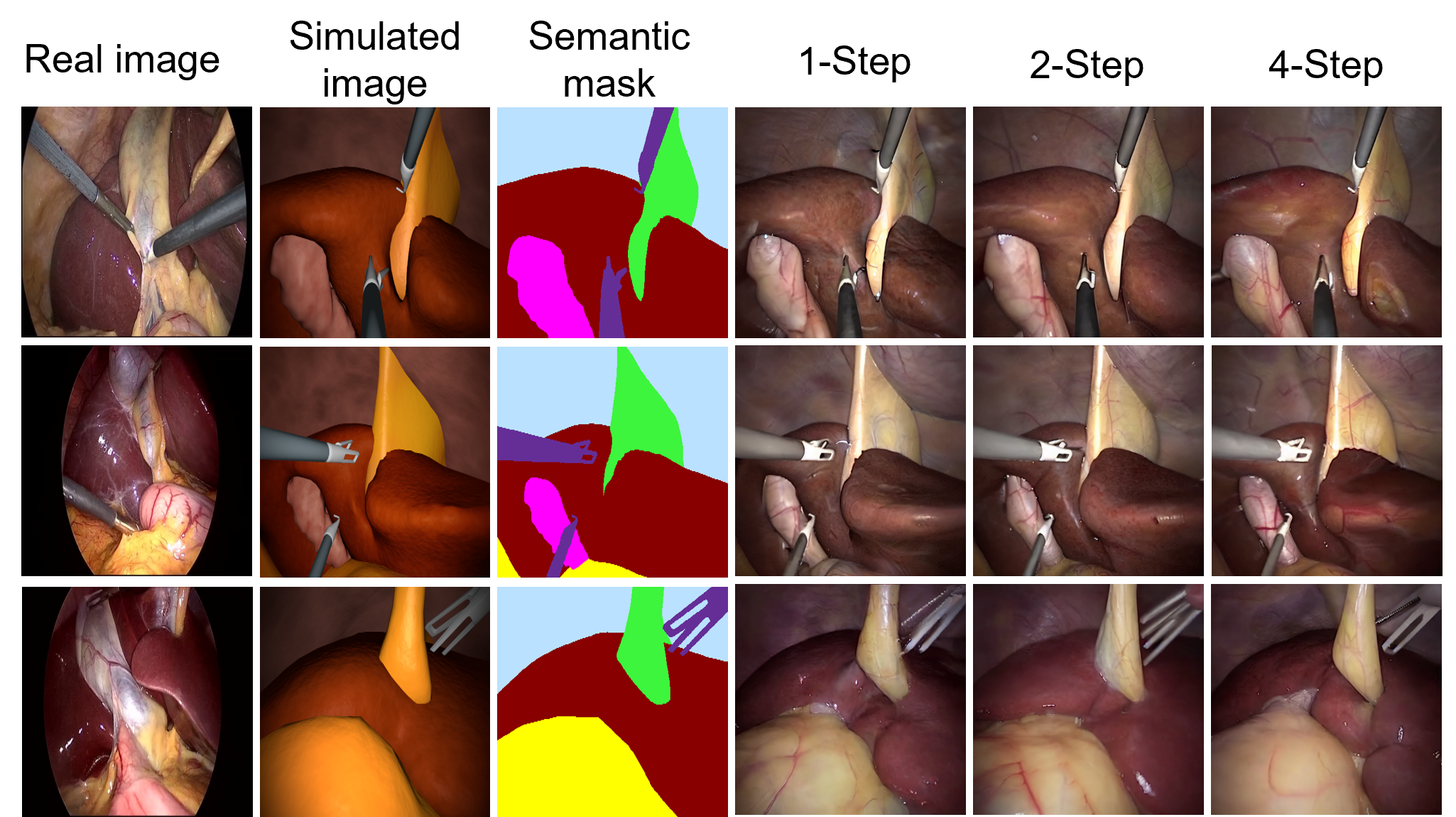}
  \caption{The realistic surgical images ($3^{rd}$-$5^{th}$ column) generated using our (few) step diffusion approach in an unpaired fashion with their corresponding semantic labels. The diffusion models are trained using the real images ($1^{st}$ column), and the simulated images ($2^{nd}$ column) are used as inputs during inference. There exists no one-on-one spatial correspondence between the real and simulated domains. Or approach is able to add fine details like vessels similar to real images.}
  \label{fig:one}
\end{figure}

In the surgical domain, simulation environments are increasingly being explored as a potential solution to this challenge~\cite{pfeiffer2019generating,yoon2022surgical,dowrick2022large}. Individual organs or abdominal cavities can be modeled in 3D environments, and synthetic surgical images can be rendered. A significant advantage of this process is that a substantial number of synthetic surgical images can be automatically generated with corresponding labels, such as semantic masks, depth, and normal maps. Nonetheless, these images lack real-world characteristics, especially the style/texture of the organs, which are different from those of real surgical datasets (See~\cref{fig:one}).

Generative modeling techniques, such as image-to-image (I2I) translation, have gained immense popularity for matching  style characteristics between domains. They have been adopted to solve multiple computer vision tasks, such as translating synthetic images into realistic ones, performing style transfer, and adapting images across different domains~\cite{benaim2017one, huang2018multimodal, liu2017unsupervised, zhang2017stackgan, choi2018stargan, hoffman2018cycada, dosovitskiy2016generating, ledig2017photo}. Importantly, for our surgical scenario, paired translation~\cite{isola2017image, liu2019few} is not possible as there exists no one-on-one correspondence between the real and synthetic images. The domains match only in the type of organs present. Hence the problem is unpaired image-to-image translation (UI2I)~\cite{zhu2017unpaired}.

In unpaired setting, generative adversarial networks (GANs)~\cite{goodfellow2020generative} have been the primary choice for translating synthetic surgical images into realistic images. Task-specific modifications to popular GAN approaches have been proposed to generate realistic surgical images or video data~\cite{pfeiffer2019generating,rivoir2021long,yoon2022surgical,sharan2021mutually,venkatesh2024exploring}. However, a domain gap persists as these methods still fail to translate the true style between domains accurately and lack the ability to generate high-quality images. This highlights the need for better generative modeling techniques to reduce the domain gap.

Diffusion models (DMs)~\cite{sohl2015deep,ho2020denoising} have established themselves as powerful generative models capable of producing high-quality, diverse images that match human perceptual levels. Text-conditioned DMs, such as Stable Diffusion (SD)~\cite{rombach2022high}, have revolutionized image generation. DMs have been increasingly used to generate medical data~\cite{kazerouni2023diffusion}, though very little work exists for surgical images. LC-SD~\cite{kaleta2024minimal} was proposed as an image translation approach using SD to generate realistic surgical images from surgical simulations. Although this approach generates high-quality images, we identified a few drawbacks. $(1).$ A pre-texturing operation was used to extract textures of different organs from real images and map them onto synthetic images before image translation using SD. To account for patient diversity, textures from each patient must be captured, making this process expensive. $(2).$ A common disadvantage of DMs is the costly image generation operation, which requires multiple iterative steps to generate a desired image~\cite{lu2022dpm,song2020denoising}. $(3).$ Finally, this work fails to evaluate the generated synthetic images to address their reliability in different surgical downstream use cases.

In this study, we aim to generate realistic surgical images from simulated images using diffusion models in an unpaired manner. Our method significantly reduces the annotation costs as the generated images have corresponding semantic labels automatically rendered from simulations. We tackle the following challenges: $(1).$ To improve the quality of the generated images, we implement a straightforward color transformation method based on optimal transport (OT)~\cite{ferradans2014regularized,perrot2016mapping} that does not require pre-texture information from the real surgical datasets. $(2).$ We minimize the number of steps necessary for image generation by employing consistency distillation in the latent space~\cite{luo2023latent}, enabling significantly faster surgical image synthesis. $(3).$ We thoroughly evaluate the generated images across various surgical downstream tasks.

We summarize our contributions as follows,
\begin{enumerate}
    \item We propose a multi-stage latent diffusion method to generate realistic surgical images with corresponding semantic labels via an unpaired image-to-image translation task. Our method requires less than five diffusion sampling steps to generate images. 
    \item We conduct experiments on three different datasets and show that our method can synthesize high-quality and diverse surgical images compared to state-of-the-art translation methods. 
    \item  We evaluate the generated surgical images on image quality, semantic consistency, and the downstream segmentation of different anatomical organs. Combining the generated images with real images indicated a performance improvement of $10\%$ across two segmentation models and three evaluation metrics.
\end{enumerate}

\section{Preliminaries}
In this section, we define the preliminary mathematical formulations necessary to introduce the diffusion process and review the consistency distillation method.
\subsection{Diffusion Models}
Diffusion models are generative models that gradually generate samples via an iterative denoising process. The data distribution $p_{data}(x)$ is added with Gaussian noise in a forward process and samples from this distribution are generated in backward process. A score-based modeling objective using stochastic differential equations (SDEs) was proposed in~\cite{song2020score} matching the probabilistic modeling of the reverse steps in DDPM~\cite{sohl2015deep,ho2020denoising}. The forward process transitions the distribution $p_{data}(x)$ to the marginal distribution $q_t(x_t)$ through a series of steps described as, $
q_{t}(x_t | x_0) = \mathcal{N}(x_t | \alpha(t)x_0, \sigma^2(t)I)$, where $\alpha(t)$ is the noise schedule. In the continuous time interval, $t\in[0,T]$, this process can be described using SDE~\cite{song2020score,lu2022dpm,karras2022elucidating}  via,
\begin{equation}
dx_t = \mu(x_t) \, dt + \eta(t) \, dw_t,
\end{equation}
where $w_t$ is a standard Brownian motion, $ \mu(\cdot) $ and \( \eta(\cdot)\)  are the drift and diffusion coefficient of \( x(t) \) respectively. To obtain a sample \( x_0 \sim p_{data}(x_0) \) starting from $x_T \sim q_T(x_T)$, a probability-flow ordinary differential equation (PF-ODE) of the reverse diffusion process is defined as
\begin{equation}\label{eq:2}
dx_t = [\mu(x_t) - \frac{1}{2}\eta^2(t)\nabla_x \log q_t(x_t)]dt,
\end{equation}
where \(\nabla \log q_t(x_t)\) is the score function which is estimated by the noise prediction model \(\epsilon_\theta(x_t, t)\), parameterized by a neural network. For classifier-free guidance (CFG)~\cite{ho2022classifier}, the original noise prediction is replaced by a linear combination of conditional and unconditional noise prediction, i.e.,

$\tilde{\epsilon}_\theta (x_t, \omega, c, t) = (1 + \omega)\epsilon_\theta (x_t, c, t) - \omega\epsilon_\theta (x_t, \emptyset, t)
$, with $\omega$ the CFG scale and $c$ a condition i.e, text.

\subsection{Consistency Models (CM)}
CM is a new family of generative models that are designed to map any point on the PF-ODE trajectory to its origin, thereby enabling a few-step generation~\cite{song2023consistency}. This is achieved through a consistency function \(h(x_t, t) \rightarrow x_\epsilon\), that satisfies a self-consistency property defined as:
$
h(x_t, t) = h(x_{t'}, t'),\forall t, t' \in [\epsilon, T]
$, indicating the outputs are consistent on any arbitrary pairs of $(x_t, t)$. The self-consistency property is enforced, and the consistency function is parameterized by a neural network \(H_\theta(x, t)\) as:

\begin{equation}
    h_\theta(x, t) = c_{\text{skip}}(t)x + c_{\text{out}}(t)H_\theta(x, t)
\end{equation}
where \(c_{\text{skip}}(t)\) and \(c_{\text{out}}(t)\) are differentiable functions. A pre-trained DM can be distilled using the consistency objective. A target model $\theta^-$ is updated i.e, \(\theta^- \leftarrow \phi \theta^- + (1 - \phi) \theta \), via exponential moving average (EMA) of the model $\theta$ that is to be learned. The consistency loss is defined as:

\begin{equation}
\mathcal{L}(\theta, \theta^-; \Phi) = \mathbb{E}_{x,t}\left[ d\left(h_\theta(x_{t_{n+1}}, t_{n+1}), h_{\theta^-}(x_{\Phi_{t_n}}, t_n) \right) \right]
\end{equation}
where \(d(\cdot, \cdot)\) is a distance metric and $\Phi$ is a one-step ODE solver. Distilling a pre-trained DM using the CM objective results in a \emph{Consistency Distillation} model.

\subsection{Latent Consistency Models}

Latent Consistency Models (LCMs) extend the concept of consistency models to the latent space of image data~\cite{luo2023latent}. Latent diffusion models (LDMs) have significantly reduced computational overhead and enhanced image generation quality~\cite{rombach2022high}. The data $x_t$ in~\cref{eq:2} is replaced by the image latents $s_t$.
The consistency function for LCMs operates in the latent space and is defined as:
\begin{equation}
h_\theta(s, c, t) = c_{\text{skip}}(t)s_t + c_{\text{out}}(t)\left( s_t - \frac{\sigma_t\epsilon_\theta(s, c, t)}{\alpha_t} \right)
\end{equation}
The loss function is given by:
\begin{equation}\label{eq:6}
\mathcal{L}(\theta, \theta^-; \Psi) = \mathbb{E}_{s, c, n}\left[ d\left( h_\theta(s_{t_{n+1}}, c, t_{n+1}), s_{\theta^-}(s_{\Psi_{t_n}}, c, t_n) \right) \right]
\end{equation}
where \( s_{\Psi_{t_n}} \) is an estimate of the evolution of the PF-ODE from \( t_{n+1} \) to \( t_n \). CFG was introduced in the form of \emph{augmented} PF-ODE in LCM to perform guided generation. Readers can refer to Section $4$ of~\cite{luo2023latent} for more details on the proof.

\subsection{Image editing with SDEdit}
SDEdit~\cite{meng2021sdedit} is a guided image editing method based on SDEs. This approach essentially noises a given user input image $x$ up to a certain noise level $x_{n}$ and denoises from this noised version $x_n$. SDEdit works on the principle that the reverse-SDE in~\cref{eq:2} can be solved not only for $t=1$ but for any $t$ in the range $(0,T)$.  A hyper-parameter $n$ denotes the extent to which the initial image is noised. Similar to DMs, a condition $c$ can be specified during the denoising process to generate the image.

\subsection{Optimal transport} Optimal transport (OT) can transform one probability distribution into another while minimizing a cost function. For color transfer, this involves mapping the color distribution of one image to another. A transport map \( T \) that minimizes the cost function \( J(\cdot) \) between the source \( \mathcal{S}_I \) and target \( \mathcal{T}_I \) distributions  was formulated as a OT by Monge~\cite{kantorovich2006translocation} as :
\begin{equation}
    \min_T \int_I J(i, T(i)) d\mathcal{S}_I(i)
\end{equation}
Assuming dicrete distributions with same number of points, the problem can be re-written as:
\begin{equation} \label{eq:t1}
    \min_{\Sigma \in P} \langle J_{\mathcal{S},\mathcal{T}}, \Sigma \rangle = \sum_{i,j=1}^N J(\mathbf{S}_i, \mathcal{T}_j) \Sigma_{i,j}
\end{equation}
where \( \Sigma \) is a permutation matrix and \( P \) is the set of permutation matrices. Relaxing~\cref{eq:t1} with the set of bi-stochastic matrices $\mathbf{B}_t$ leads to partial transport by
\begin{equation}\label{eq:t2}
    \min_{\Sigma \in \mathcal{B}_t} \langle J_{\mathcal{S},\mathcal{T}}, \Sigma \rangle
\end{equation}
For the color transfer problem, strict mass conservation is not necessary. To match the point clouds (pixels) between the two images~\cref{eq:t2} is further relaxed via :
\begin{equation} 
\min_{\Sigma \in \mathcal{B}_\kappa} \langle \Sigma, J_{\mathcal{S},\mathcal{T}} \rangle + \lambda_\mathcal{S} J_{p,q} (G_\mathcal{S} \Delta_{\mathcal{S},\mathcal{T}} (\Sigma)) + \lambda_\mathcal{T} J_{p,q} (G_\mathcal{T} \Delta_{\mathcal{T},\mathcal{S}} (\Sigma^*))
\end{equation}

where \( \lambda \) is the regularization parameter, set to $1$ in this work, \( J_{p,q} \) is a norm, and \( G_{(\cdot)} \) is the gradient operator on the graph structure of the point clouds. For more details on the proof, readers can refer to~\cite{ferradans2014regularized}.

\section{Methods}
\begin{figure}[tb]
  \centering
  \includegraphics[height=6.5cm]{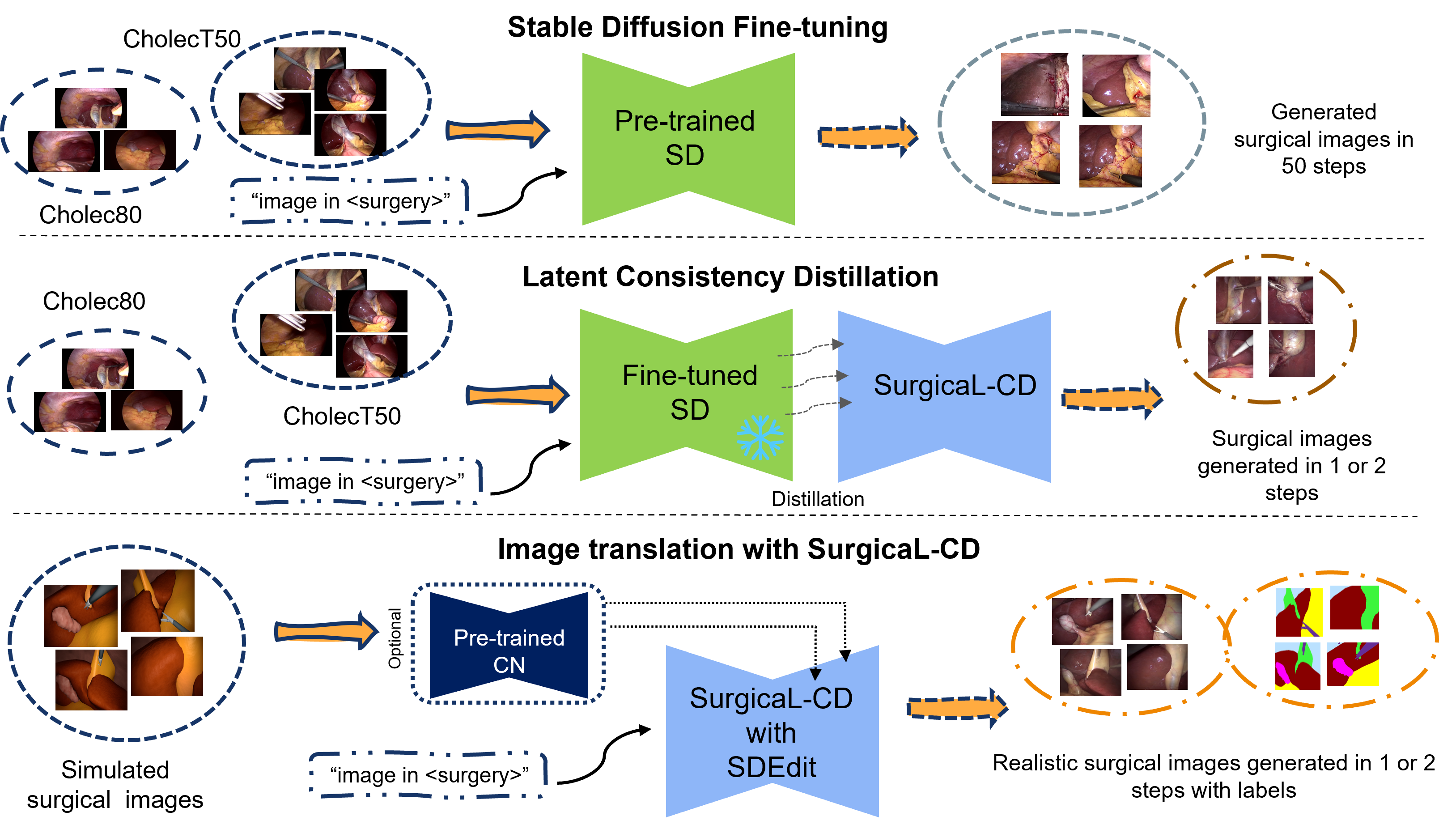}
  \caption{\textbf{Our unpaired few-step diffusion method}. As the $1^{st}$ stage, SD~\cite{rombach2022high} model is fine-tuned using the text prompts on each real surgical dataset. After training, this model is capable of generating surgical images. In the $2^{nd}$ stage, consistency distillation of the fine-tuned model occurs using real surgical images and text prompts. We call this model \emph{SurgicaL-CD} that generates surgical images in a few steps given a text prompt. Finally, the simulated images are given as input to the \emph{SurgicaL-CD} model, which uses SDEdit~\cite{meng2021sdedit} for image translation. In this manner, the simulated images are translated into realistic surgical images. To preserve the structure of different organs, pre-trained ControlNet~\cite{zhang2023adding} is optionally used in the inference pipeline. }
  \label{fig:method}
\end{figure}

In this section, we briefly describe the individual components of our diffusion pipeline. Our method consists of the Stable Diffusion model that is consistently distilled to generate surgical images in a few steps. The SDEdit method is combined with distilled SD model to generate realistic surgical images with simulated images as input.
 An overview of our approach is shown in~\cref{fig:method}.
\subsection{Stable Diffusion fine-tuning}
Stable Diffusion (SD)~\cite{rombach2022high} is an LDM, which performs the diffusion process in the latent space in contrast to image pixel space. An autoencoder, consisting of an encoder \( E \) and a decoder \( D \), is initially trained to compress high-dimensional image data into a low-dimensional latent vector \( s = E(x) \). This latent vector is then decoded to reconstruct the image as \( \hat{x} = D(s) \). Training diffusion models within the latent space significantly reduces computational costs compared to pixel-based models and accelerates the inference process. SD model is trained on the large corpus of image-text pairs on the LAION dataset~\cite{schuhmann2022laion}. Firstly, we fine-tune this pre-trained model on the surgical datasets with corresponding text prompts. After this step, the fine-tuned model is capable of generating surgical images.
\subsection{Latent consistency distillation of SD}
In the second stage, we perform latent consistency distillation of the fine-tuned SD model using the objective in~\cref{eq:6}. This step leads to \textbf{Surgica}l \textbf{L}atent \textbf{C}onsistency \textbf{D}istill (\emph{SurgicaL-CD}) model, which is capable of generating high-quality surgical images using text prompts in only a few steps. Based on the previous method~\cite{luo2023latent}, we limit the number of steps to four in this work.   

\subsection{Image translation}
We perform a color adaptation between the domains using OT. This step is optional and depends upon the color differences between the simulated and real domains. We chose the OT-mapping from~\cite{perrot2016mapping} to adapt the pixel colors between the domains. The \emph{SurgicaL-CD} model is combined at this stage with SDEdit~\cite{meng2021sdedit} to perform the unpaired image-to-image translation. We also use the text prompts as additional controls to enforce the style/texture aspects of the real datasets to the simulated images. 

\subsection{Spatial conditioning with ControlNets}
ControlNets (CN)~\cite{zhang2023adding} are models that have been used to control the spatial alignment of different objects in SD. As an additional option, we plug the ControlNet into the SurgicaL-CD model during the image translation stage. This leads to the generation of precisely controlled anatomical structures with the texture properties of real images. Training CNs from scratch or even fine-tuning them requires large amounts of data (refer to suppl. material of~\cite{zhang2023adding}). Hence, we adopted two pre-trained models the soft edge and depth-controlled CNs as zero-shot solution in this work.

\section{Experiments}
We outline our experiments in this section, where we compare and evaluate the generated images from our approach on image quality, semantic consistency, and downstream utility as training data.
\subsection{Data}
\textbf{Real surgical datasets}. We use the CholecSeg8K~\cite{hong2020cholecseg8k} dataset, which consists of $8080$ images of cholecystectomy. The dataset is a labeled subset of Cholec80~\cite{endonet} and contains the abdominal wall, liver, gall bladder, liver ligaments, fat, and surgical tools. Following~\cite{kaleta2024minimal}, we also use the CholecT50~\cite{nwoye2022rendezvous} dataset (cholecystectomy) that contains five additional patients to Cholec80. 
As a third dataset, we use the multi-class subset of Dresden Surgical Anatomy dataset (DSAD)~\cite{carstens2023dresden}consisting of $1400$ images. The dataset consists of the following organs: abdominal wall, liver, colon, stomach, pancreas, and small intestine. We use the official dataset splits as suggested in~\cite{nwoye2022rendezvous,kolbinger2023anatomy} for training and testing our models.

\textbf{Surgical simulation datasets}. We use $10,000$ surgical simulated scenes from Pfeiffer\etal~\cite{pfeiffer2019generating} (\emph{Lap scenes}) to generate the translated images. Similarly $4000$ surgical scenes from Yoon\etal~\cite{yoon2022surgical} (\emph{Gast scenes}) were used. 
The simulated scenes and the real images should match in the type of the organs present in them. Hence, we use the \emph{Lap scenes} to generate images with Cholec80 and CholecT50 datasets and \emph{Gast scenes} with the DSAD dataset. 

\subsection{Evaluation}
\textbf{Image quality}.
We employ different metrics to compare the generated image quality. We use the popular metric  CFID~\cite{parmar2022aliased} 
to measure the feature distributions between the real and generated images from pre-trained networks. The image quality on features extracted from CLIP is used to compute CMMD~\cite{jayasumana2024cmmd} score and is suitable for smaller datasets as it is unbiased, unlike CFID. Density and coverage (D\&C)~\cite{naeem2020reliable}  assess the image quality and diversity respectively only based on image embedding and k-nearest neighbor manifolds and do not rely on pre-trained models, which can differ in modality to surgical images. We use the improved implementation of D\&C~\cite{venkatesh2022detecting}. Finally, we compute the LPIPS~\cite{zhang2018unreasonable} metric that measures the perceptual quality of the images. As unpaired GAN-based baselines, we chose surgical focussed methods such as LapMUNIT~\cite{pfeiffer2019generating} and ConStructS~\cite{venkatesh2024exploring} and along with CUT~\cite{park2020contrastive} and CycleGAN~\cite{zhu2017unpaired}. The SDEdit~\cite{meng2021sdedit}, LC-SD~\cite{kaleta2024minimal} and CycleDiffusion~\cite{wu2023latent} served as diffusion-based baselines. We added SDEdit~\cite{meng2021sdedit} with the fast solver, DPM++-Solver~\cite{lu2022dpm} as a qualitative baseline. We used the generated images from LC-SD~\cite{kaleta2024minimal}, which uses pre-texturing operation before translation. All the other baselines were trained on the Cholec80, CholecT50 and DSAD datasets from scratch.

\textbf{Semantic consistency assessment}.
We compute the metrics based on off-the-shelf segmentation models to assess the semantic consistency (i.e., label preservation)~\cite{venkatesh2024exploring,kaleta2024minimal} during translation. A segmentation model is trained on the real images of the specific dataset. Using the ground truth segmentation masks we can measure the segmentation accuracy of the translated images. This approach is based on the premise that both realism and preserved semantic content are essential for achieving high segmentation scores~\cite{isola2017image}. To assess the performance difference between the Cholec80 and CholecT50 datasets, we train our method and the baselines on them separately. We train the segmentation model on the CholecSeg8k dataset and use this pre-trained model to compute the metrics. 

\textbf{Downstream-semantic segmentation}.
We evaluate the utility of the generated synthetic images in the downstream segmentation of real datasets. This evaluation emphasizes the importance of generating surgical images from simulated images. We train segmentation models on the translated synthetic images and then fine-tune them on the real images. We assess these datasets' performance on multi-class segmentation using the Dice, IOU, and Hausdorff distance as metrics following~\cite{maier2024metrics}. We train DV3+~\cite{chen2018encoder}, Segformer~\cite{xie2021segformer}, and Upernet~\cite{xiao2018unified} segmentation models to assess the performance across different model capacities and architectures. As baselines, we train the models with no augmentations, color, and the combination of color and spatial augmentations for each model.\\

\textbf{Implementation details}. We use Stable Diffusion~\cite{rombach2022high} v$1.5$ as the base model. We fine-tune this model using Adam optimizer~\cite{kingma2017adammethodstochasticoptimization} with a learning rate of $1e^{-5}$. We fine-tune the SD on the three datasets separately using text prompts based on the datasets i.e, \emph{an image of cholect50}, for the CholecT50 dataset during training and inference. We used the DDIMSolver~\cite{song2020denoising} with DDPM~\cite{ho2020denoising} scheduler for the consistency distillation process. The CFG scales in the range $4.5-7.5$ were used. We add a denoising strength of $0.5$ for all our models. To maintain the structure of the organs, we used ControlNet pre-trained on soft edges and also depth maps. 
\begin{table}[t]
  \caption{\textbf{Image quality assessment against baselines on Cholec80.} CFID and LPIPS measures the realism of the images, while Density and Coverage indicates perceptual quality and diversity of the generated images. SDEdit~\cite{meng2021sdedit} shows the best CFID scores, whereas our method clearly outperforms the GAN and diffusion approaches on all the other metrics. $st$ denotes the noise strength. Best scores are denoted in \textbf{bold}.}
  \label{tab:comp_img}
  {\small{
    \resizebox{\linewidth}{!}{
\begin{tabular}{llccccc}
\toprule
    Approach & Method & CFID ($\downarrow$) & Density ($\uparrow$) & Coverage ($\uparrow$)& CMMD($\downarrow$) & LPIPS($\downarrow$) \\
\midrule
& CUT~\cite{park2020contrastive} & $147.73$& $0.01$& $1.1$& $2.652$& $0.668$\\
\multirow{4}{1cm}{\emph{GAN} based} & CycleGAN~\cite{zhu2017unpaired}  & $182.27$& $0.06$& $2.3$& $1.998$& $0.652$\\
 &  LapMUNIT~\cite{pfeiffer2019generating} & $217.89$& $0.11$& $2.1$& $5.138$& $0.671$\\
& ConStructS~\cite{venkatesh2024exploring} & $145.26$& $0.06$& $2.3$& $2.245$& $0.642$\\ \hdashline
& LC-SD~\cite{kaleta2024minimal} & $142.28$& $0.05$& $1.4$& $2.609$& $0.707$\\
 \multirow{5}{1cm}{\emph{Diffusion} based}& CycleDiffusion~\cite{wu2023latent} & $210.07$ & $0.04$ & $0.09$ & $4.023$ & $0.765$ \\
& SDEdit~\cite{meng2021sdedit}($st=0.25$)  & $205.37$& $0.15$& $2.2$& $3.590$& $0.675$\\
 & SDEdit~\cite{meng2021sdedit}($st=0.50$)  & $187.49$& $0.15$& $2.4$& $1.706$&$0.681$ \\
& SDEdit~\cite{meng2021sdedit}($st=0.75$)  & $\mathbf{138.55}$&$0.13$ &$2.2$ &$1.662$ &$0.682$ \\ \hdashline
\multirow{3}{1cm}{\emph{Ours}}& $1$-Step & $203.23$&$0.17$ &$\mathbf{2.8}$ &$\mathbf{1.565}$ &$\mathbf{0.602}$ \\
 &$2$-Step &$195.21$ &$\mathbf{0.19}$ &$2.7$ &$1.660$ &$0.617$ \\
& $4$-Step &$173.90$ &$0.15$ &$2.5$ &$1.650$ &$0.620$ \\
    \bottomrule
\end{tabular}
}
}}
\end{table}

\subsection{Results}
\textbf{Cholecystectomy datasets}. ~\cref{tab:comp_img} presents the image quality results of the different image translation methods. SDEdit~\cite{meng2021sdedit}, with a denoising strength of 0.75, achieves the best CFID score but falls short compared to our method in all other metrics. Our $1$-step and $2$-step methods generate high-quality and diverse images, as indicated by the high density and coverage values. Similarly, our method demonstrates overall lower LPIPS scores compared to other diffusion approaches, indicating its capability to capture the texture properties of real images during image translation.

\begin{figure}
  \centering
  \includegraphics[width=12cm]{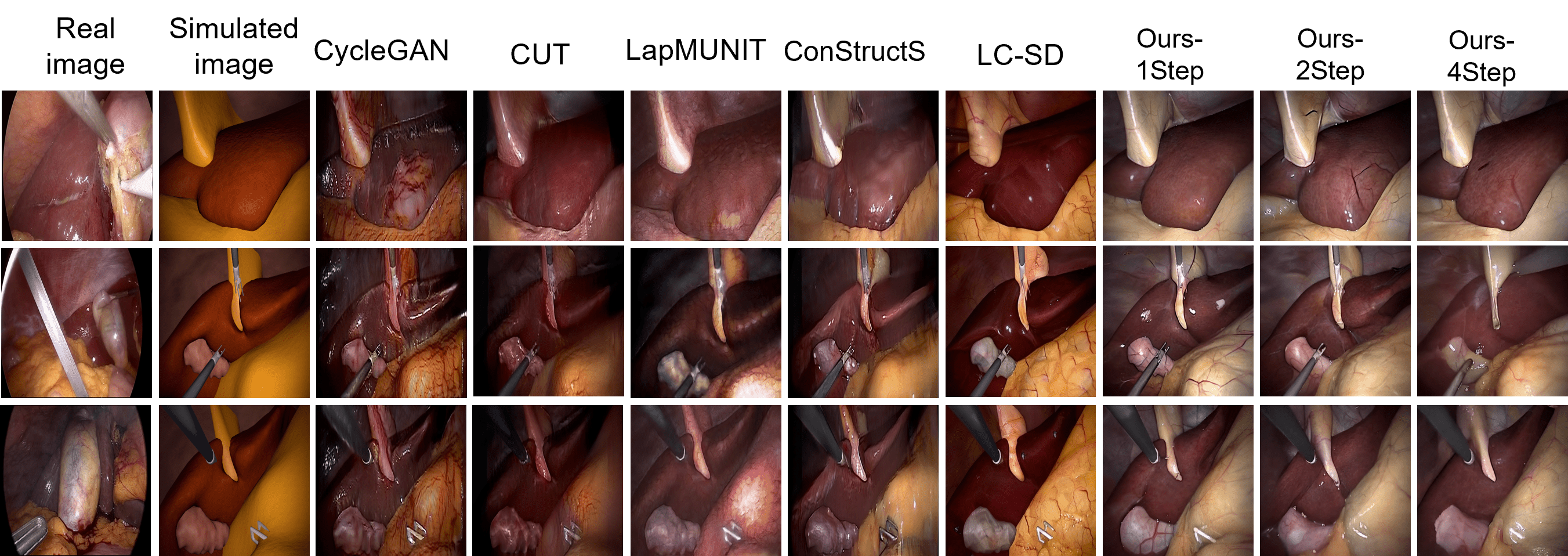}
  \caption{\textbf{Comparison to baselines on the Cholec80 dataset.} The translated images from our method are compared to state-of-the-art GAN and diffusion methods. Our method can add finer details like vessels and generate the real textures of the organs directly during translation. The pre-texturing patterns are visible after translation in LC-SD~\cite{kaleta2024minimal}, while GAN approaches lack image quality.}
  \label{fig:comp}
\end{figure}

\begin{table}
\caption{\textbf{The semantic quality comparison between baselines.} Inference time denotes the time taken to generate an image from the model. A segmentation model pre-trained on real surgical images is used to infer the generated images against their ground truth semantic label. Our method outperforms the previous GAN baselines. For the CholecT50 dataset, our method matches SDEdit~\cite{meng2021sdedit} scores and even outperforms it on the Cholec80 dataset, having $10\times$ faster inference time. The best scores are denoted in \textbf{bold}.}
   {\small{
    \resizebox{\textwidth}{!}{
\begin{tabular}{lp{1cm}cccccc}
\toprule
    Method & Inference time & \multicolumn{3}{c}{Cholec80~\cite{hong2020cholecseg8k}} & \multicolumn{3}{c}{CholecT50~\cite{nwoye2022rendezvous}} \\
    \cmidrule(lr){3-5} \cmidrule(lr){6-8}
     & & pxAcc($\uparrow$) & IOU($\uparrow$) & Dice($\uparrow$) & pxAcc($\uparrow$) & IOU($\uparrow$) & Dice($\uparrow$)\\
\hline
Simulated images & - & $0.28$ & $0.13$ & $0.19$ & $-$ & $-$ & $-$\\
CUT~\cite{park2020contrastive} & $0.11$ s& $0.45$ & $0.29$ & $0.42$ & $0.49$ & $0.31$ & $0.40$\\
CycleGAN~\cite{zhu2017unpaired}  & $0.04$ s & $0.42$ & $0.30$ & $0.42$ & $0.46$ & $0.28$&$0.41$ \\
LapMUNIT~\cite{pfeiffer2019generating} & $0.03$ s& $0.41$ & $0.25$ & $0.37$ & $0.39$&$0.24$ &$0.37$ \\
ConStructS~\cite{venkatesh2024exploring} & $0.11$ s & $0.47$ & $0.30$ & $0.43$ & $0.50$&$0.31$ &$0.43$ \\ \hdashline
LC-SD~\cite{kaleta2024minimal} & $5.52$ s & $0.42$&$0.26$ &$0.39$ &$0.41$ &$0.26$ &$0.38$ \\
CycleDiffusion~\cite{wu2023latent} & $5.59$ s & $0.32$ & $0.21$ & $0.20$ & $0.30$ & $0.14$& $0.18$\\
SDEdit~\cite{meng2021sdedit}($st=0.25$) & $1.13$ s & $0.50$ & $0.32$ & $0.41$ & $\mathbf{0.52}$& $\mathbf{0.34}$& $\mathbf{0.47}$\\
SDEdit~\cite{meng2021sdedit}($st=0.50$) & $1.76$ s & $0.50$ & $0.34$ & $0.43$ & $0.50$& $0.33$& $0.43$\\
SDEdit~\cite{meng2021sdedit}($st=0.75$) & $2.31$ s & $0.45$ & $0.29$ & $0.42$ & $0.45$ & $0.29$ & $0.42$ \\ \hdashline
Ours ($1$ Step) & $0.14$ s & $\mathbf{0.52}$ & $\mathbf{0.35}$& $\mathbf{0.47}$& $0.51$ &$\mathbf{0.34}$ & $0.46$\\
Ours ($2$ Step) & $0.17$ s & $0.50$ & $0.33$& $0.46$& $\mathbf{0.52}$ &$\mathbf{0.34}$ & $\mathbf{0.47}$\\
Ours ($4$ Step) & $0.23$ s & $0.49$ & $0.32$ & $0.45$ & $0.50$ & $0.33$ & $0.46$ \\
    \bottomrule
\end{tabular}
}
}}
\label{tab:comp_label}
\end{table}

The qualitative results for the Cholec80 dataset are shown in~\cref{fig:comp}. GAN-based approaches maintain the anatomical structures well but fail to transfer the textures/style from real images. CycleGAN~\cite{zhu2017unpaired} introduces improper textural patterns, degrading image quality. The pre-texturing patterns in LC-SD~\cite{kaleta2024minimal} are quite visible, indicating its inability to translate textures from real images effectively. Our method successfully adds finer details, such as vessels, directly during translation. The organ structures are slightly distorted with the 4-step method; however, this can be improved with better CFG and denoising strength scales. We leave this exploration for future work.

\begin{figure}[t]
  \centering
  \includegraphics[width=12cm]{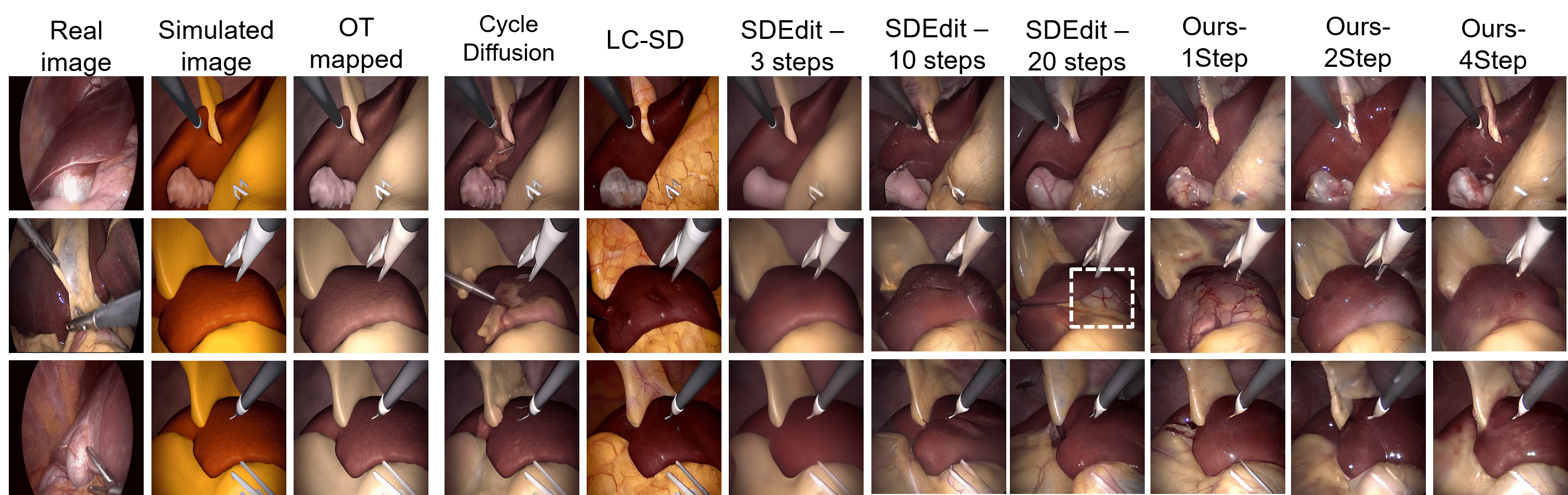}
  \caption{\textbf{Comparison to baselines on the CholecT50 dataset.} The $3^{rd}$ column shows the simulated images after OT mapping. LC-SD~\cite{kaleta2024minimal} fails to adapt the color and texture of real images. Texture transfer does not occur using the DPM++-solver~\cite{lu2022dpm} with SDEdit~\cite{meng2021sdedit} for $3$ or $10$ steps. $20$-Step SDEdit shows good image quality, while the absence of structure control leads to a hallucinated gall bladder ($2nd$ row)(white box). Our method ($1$ and $2$ step) can maintain the shape and transfer fine texture details from real images.}
  \label{fig:comp_t50}
\end{figure}

\begin{table}[t]
\begin{center}
  \caption{\textbf{The downstream segmentation scores on Cholec datasets.} The first three rows indicate the performance using only real images. The translated synthetic images are combined with the real images (+Real) for training. Across both architectures, the combined training method shows the best performance. The best scores for Cholec80 and CholecT50 are shown in \textbf{bold}.}
  \label{tab:c80_comp}
   {\small{
    \resizebox{0.9\textwidth}{!}{
\begin{tabular}{lcccccc}
\toprule
    Training scheme &  \multicolumn{3}{c}{DV3+~\cite{chen2018encoder}} & \multicolumn{3}{c}{UperNet~\cite{xiao2018unified}} \\
    \cmidrule(lr){2-4} \cmidrule(lr){5-7}
     & Dice($\uparrow$) & IOU($\uparrow$) & HD($\downarrow$) & Dice($\uparrow$) & IOU($\uparrow$) & HD($\downarrow$)\\
\midrule
No-aug  & $0.50$ & $0.36$ & $115.36$ & $0.56$ & $0.47$ & $118.37$\\
Color-aug & $0.53$ & $0.39$ & $\mathbf{101.54}$ & $0.59$ & $0.45$ & $110.93$\\
Color+spatial-aug & $0.58$ & $0.45$ & $108.14$ & $0.61$ & $0.50$ & $108.63$\\ 
\midrule
\multicolumn{7}{c}{SurgicaL-CD trained on Cholec80~\cite{hong2020cholecseg8k}} \\
\midrule
Ours ($1$ Step)+Real & $\mathbf{0.67}$ & $\mathbf{0.53}$& $110.41$&$\mathbf{0.69}$ & $\mathbf{0.56}$ & $109.72$\\
Ours ($2$ Step)+Real  & $0.66$ & $0.52$& $109.12$& $0.66$ &$0.53$ & $\mathbf{107.60}$\\
Ours ($4$ Step)+Real  & $0.59$ & $0.49$ & $110.34$ & $0.67$ & $0.54$ & $109.65$ \\
\midrule
\multicolumn{7}{c}{SurgicaL-CD trained on CholecT50~\cite{nwoye2022rendezvous}} \\
\midrule
Ours ($1$ Step)+Real & $0.58$ & $0.46$& $109.32$ &$\mathbf{0.67}$ & $\mathbf{0.54}$ & $109.72$\\
Ours ($2$ Step)+Real  & $\mathbf{0.68}$ & $\mathbf{0.55}$& $\mathbf{100.12}$& $0.62$ &$0.48$ & $\mathbf{108.56}$\\
Ours ($4$ Step)+Real  & $0.60$ & $0.45$ & $111.28$ & $0.60$ & $0.46$ & $108.97$ \\
    \bottomrule
\end{tabular}
}
}}
\end{center}
\end{table}
The results of the semantic consistency evaluation are presented in~\cref{tab:comp_label}. ConStructS~\cite{venkatesh2024exploring} and CycleGAN~\cite{zhu2017unpaired} exhibit comparable performance among GAN-based methods. SDEdit~\cite{meng2021sdedit}, with a strength of 0.25, outperforms the LC-SD~\cite{kaleta2024minimal} method on the Cholec80 dataset. Our $1$-step method achieves a dice score of 0.35 and an IOU of 0.47, ranking highest among the models for the Cholec80 dataset. Minor improvements in scores are observed for the CholecT50 dataset among the GAN baselines. Despite using pre-textures from the CholecT50 dataset, LC-SD~\cite{kaleta2024minimal} still lags behind our method. Our $2$-step method surpasses the baselines by approximately $4$ points across all metrics. In terms of inference speed, our method exceeds the diffusion baselines and is comparable to GAN approaches while demonstrating superior performance overall. We did not observe significant differences in the performance of our methods between the Cholec80 and CholecT50 datasets.

\begin{table}[htb!]
  \begin{center}
  \caption{\textbf{The segmentation results on the DSAD dataset}. Combining the translated images with real images for training leads to $5\%$ improvement in dice, while larger gains are noticed in HD scores for the Segformer models.}
  \label{tab:dsad}
   {\small{
    \resizebox{0.85\textwidth}{!}{
\begin{tabular}{lcccccc}
\toprule
    Training scheme &  \multicolumn{3}{c}{DV3+~\cite{chen2018encoder}} & \multicolumn{3}{c}{Segformer~\cite{xie2021segformer}} \\
    \cmidrule(lr){2-4} \cmidrule(lr){5-7}
     & Dice($\uparrow$) & IOU($\uparrow$) & HD($\downarrow$) & Dice($\uparrow$) & IOU($\uparrow$) & HD($\downarrow$)\\
\midrule
No-aug  & $0.73$ & $0.62$ & $137.65$ & $0.78$ & $0.67$ & $98.32$\\
Color-aug & $0.76$ & $0.65$ & $66.86$ & $0.79$ & $0.69$ & $54.82$\\
Color+spatial-aug & $0.74$ & $0.64$ & $\mathbf{65.83}$ & $0.80$ & $0.70$ & $72.58$\\ \hdashline
Ours ($1$ Step+Real) & $0.77$ & $0.68$& $88.67$& $0.80$ & $0.72$ & $\mathbf{43.47}$\\
Ours ($2$ Step+Real)  & $\mathbf{0.78}$ & $\mathbf{0.69}$& $124.01$& $\mathbf{0.83}$ &$\mathbf{0.76}$ & $63.54$\\
Ours ($4$ Step+Real)  & $0.70$ & $0.59$ & $101.45$ & $0.78$ & $0.68$ & $98.32$ \\
    \bottomrule
\end{tabular}
}
}}
\end{center}
\end{table}

\begin{figure}[t]
\centering
  \includegraphics[height=6.5cm]{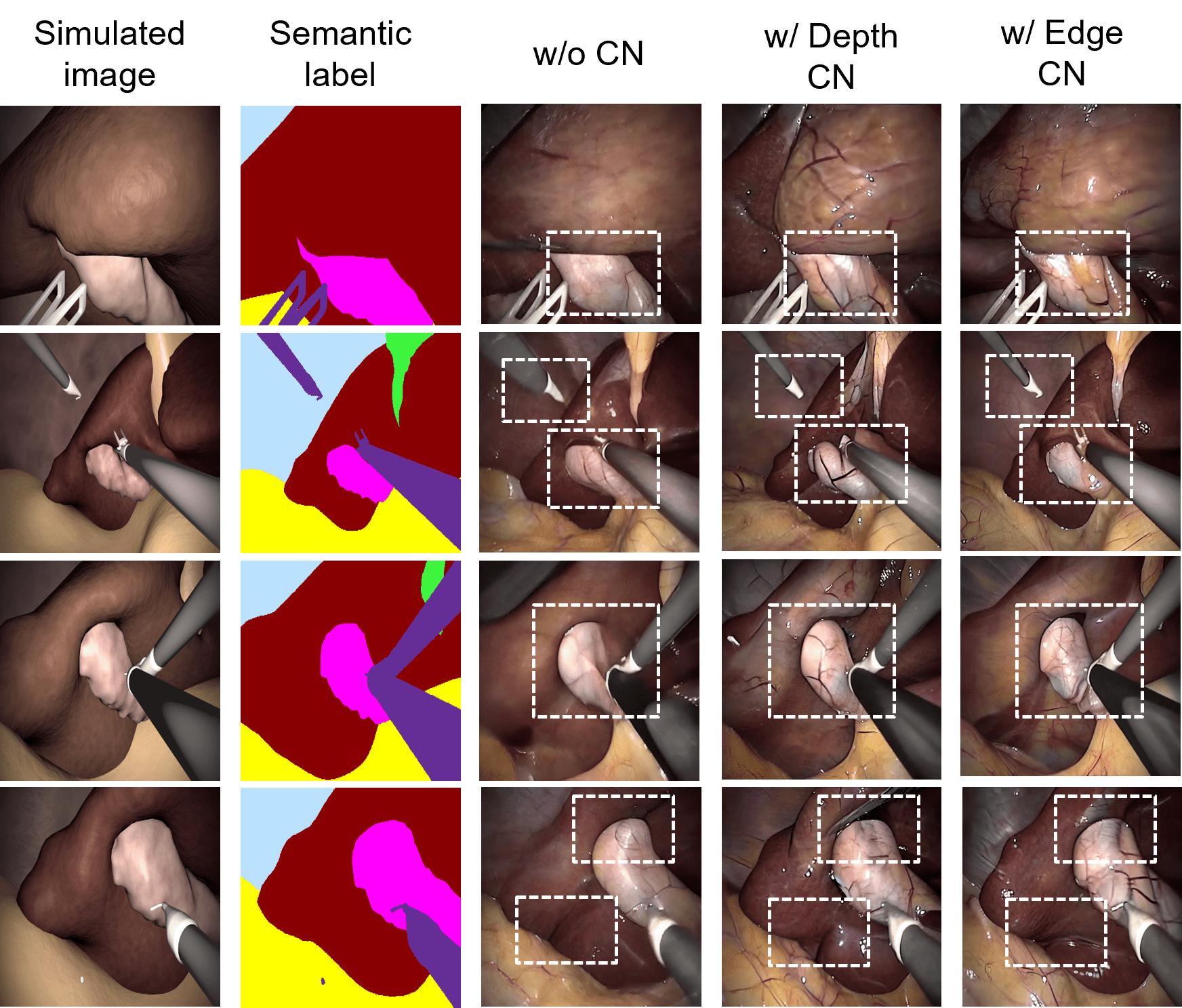}
  \caption{\textbf{Ablation results on our $1$-step method.} Without the ControlNet~\cite{zhang2023adding}, the edges of the organs and surgical tools are smoothened (indicated with white boxes). Semantic style leakage occurs in depth CNs, whereas edge control can maintain the structure and style of the generated images similar to the real images. }
  \label{fig:cn}
\end{figure}

~\cref{fig:comp_t50} shows the qualitative results on the CholecT50 dataset. CycleDiffusion~\cite{wu2023latent} causes semantic leakage during image translation. Similarly, LC-SD~\cite{kaleta2024minimal} lacks in adapting the true texture of the real images to the simulated images. The texture transfer does not occur for SDEdit within three steps. However, increasing the number of steps to $20$ leads to SDEdit hallucinating organs, indicating the importance of spatial control. Our method ($1$- and $2$-step) generates realistic surgical images while maintaining the organ structures precisely.

The downstream semantic segmentation results are shown in~\cref{tab:c80_comp}. The results indicate that combined trained with translated images and real images leads to an improvement of $9\%$ in dice and $8\%$ in IOU for the DV3+ model. Similar gains were also noticed for the UperNet architecture. While using the CholecT50 dataset, we noticed similar improvements in the segmentation scores. These results further showcase that our generated images can serve effectively as training data for segmentation. 

The segmentation results on the DSAD dataset is shown in~\cref{tab:dsad}. We notice that adding the translated images to the real images completely outperforms other training methods across both the models. Improvements ranging from $5\%-10\%$ was noticed with the combined training scheme.

\textbf{Ablation study on ControlNets.} In~\cref{fig:cn}, we show the qualitative results from the ablation study on our method, comparing methods with and without spatial conditioning using ControlNets. Combining the ControlNet enhances the preservation of organ structure and texture transfer during image translation. However, a higher conditioning scale from these pre-trained models can alter the texture of specific organs. This trade-off can be managed by adjusting the CFG scales.

\section{Discussion and Limitations}
In this work, we introduce a diffusion based unpaired image translation approach to generate high-quality, realistic surgical images from simulated images. Our approach generates images in few sampling steps and does not require any real labeled data, thereby drastically reducing the annotation costs. We thoroughly evaluated the generated images via different methods and showcased their utility in downstream applications, highlighting the importance of surgical image generation. By incorporating temporal components, our method can be used to train/generate surgical videos with very minimal overhead.

\textbf{Limitations}. Although our work shows that high-quality surgical images can be generated in a few steps, it does have certain limitations. An analysis of various CFG scales in conjunction with denoising strength is necessary to further enhance image quality. Additionally, exploring GAN-based objectives or style injection via cross-attention could help mitigate semantic (texture) leakage between organs. Given the minimal sampling steps required, implementing such techniques would incur minimal overhead. \\

\textbf{Acknowledgements}. This work is partly supported by BMBF (Federal Ministry of Education and Research) in DAAD project 57616814 (\href{https://secai.org/}{SECAI, School of Embedded Composite AI}). This work is partly funded by the German Research Foundation (DFG, Deutsche Forschungsgemeinschaft) as part of Germany’s Excellence Strategy – EXC 2050/1 –Project ID 390696704 – Cluster of Excellence “Centre for Tactile Internet with Human-in-the-Loop” (CeTI) of Technische Universit¨at Dresden.

%
%
\bibliographystyle{splncs04}
\bibliography{main}
\end{document}